# Conflict-based Force Aggregation

**John Cantwell**[1]
Department of Infrastructure and Planning
Royal Institute of Technology
SE-100 44 Stockholm, Sweden

cantwell@infra.kth.se

**Johan Schubert**[2] and **Johan Walter**
Department of Data and Information Fusion
Division of Command and Control Warfare Technology
Swedish Defence Research Agency
SE–172 90 Stockholm, Sweden
schubert@foi.se johanw@foi.se

## Abstract

In this paper we present an application where we put together two methods for clustering and classification into a force aggregation method. Both methods are based on conflicts between elements. These methods work with different type of elements (intelligence reports, vehicles, military units) on different hierarchical levels using specific conflict assessment methods on each level. We use Dempster-Shafer theory for conflict calculation between elements, Dempster-Shafer clustering for clustering these elements, and templates for classification. The result of these processes is a complete force aggregation on all levels handled.

## 1 Introduction

In this application oriented paper we report the first result from an ongoing project at the Swedish Defence Research Agency investigating which automatic conclusions can be drawn about force deployment based on low level intelligence. The scenario studied is an army scenario but in a future situation with a high flow of automatically generated intelligence reports concerning vehicles.

Our idea is to investigate the intelligence reports through a series of aggregation processes and draw automatic conclusions about force strength, deployment and ongoing actions to establish a solid basis for enemy prediction and decision superiority. In the longer run our aim is to automate the Intelligence Preparation of the Battlefield (IPB) process [18].

In section 2 of this paper we give a detailed problem description, followed by an overview in section 3 of an actual scenario. In section 4 we describe an aggregation procedure in three major steps, aggregating the information upwards level by level: first, we investigate methods for assessing conflict (using Dempster's rule [16]) between elements, i.e., intelligence reports, vehicles or units, depending on level. Secondly, we use these conflicts to partition the elements into groups (through Dempster-Shafer clustering [1–2, 5–14]) forming elements one level up in the hierarchy. Finally, we use templates and multiple hypothesis evaluation to classify grouped elements into an element on the higher level. This process is carried on level by level as long as data permits, Figure 1. Results are discussed in section 5 and conclusions are drawn in section 6.

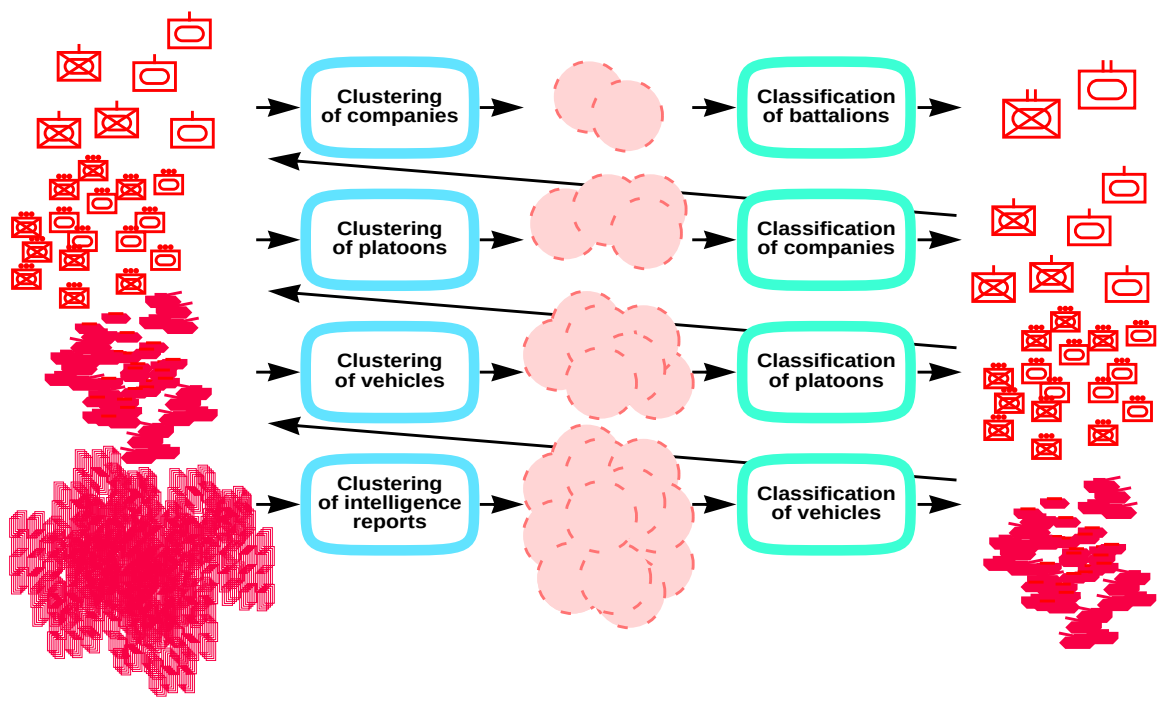

Figure 1: The aggregation process hierarchy.

1. This work was done while the author was with the Swedish Defence Research Agency.

2. http://www.foi.se/fusion/

## 2 Problem description

### 2.1 *Unit model hierarchy*

Our unit model consists of a mechanized company. In this model, the company consists of a company commander, three mechanized platoons and a main battle tank (MBT) platoon or an anti-tank missile platoon.

A mechanized platoon consists of four armored personnel carriers (tracked), a main battle tank (MBT) platoon consists of five MBTs and an anti-tank missile platoon consists of five anti-tank missile launchers (see Figure 2).

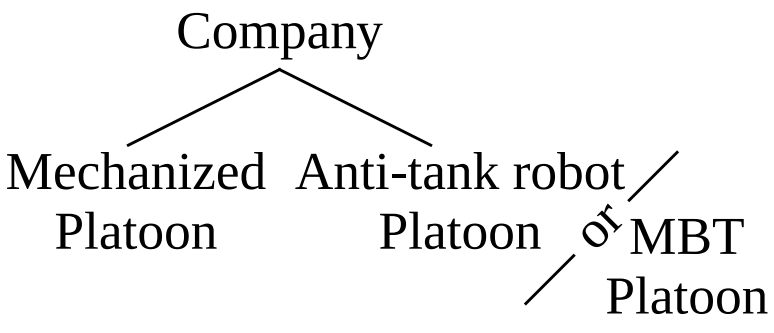

Figure 2: The hierarchy of units.

### 2.2 *Environment*

To generate sensor reports, we use the *FbSim* simulator. FbSim has been developed to analyze military units in combat situations[1] (see Figure 3).

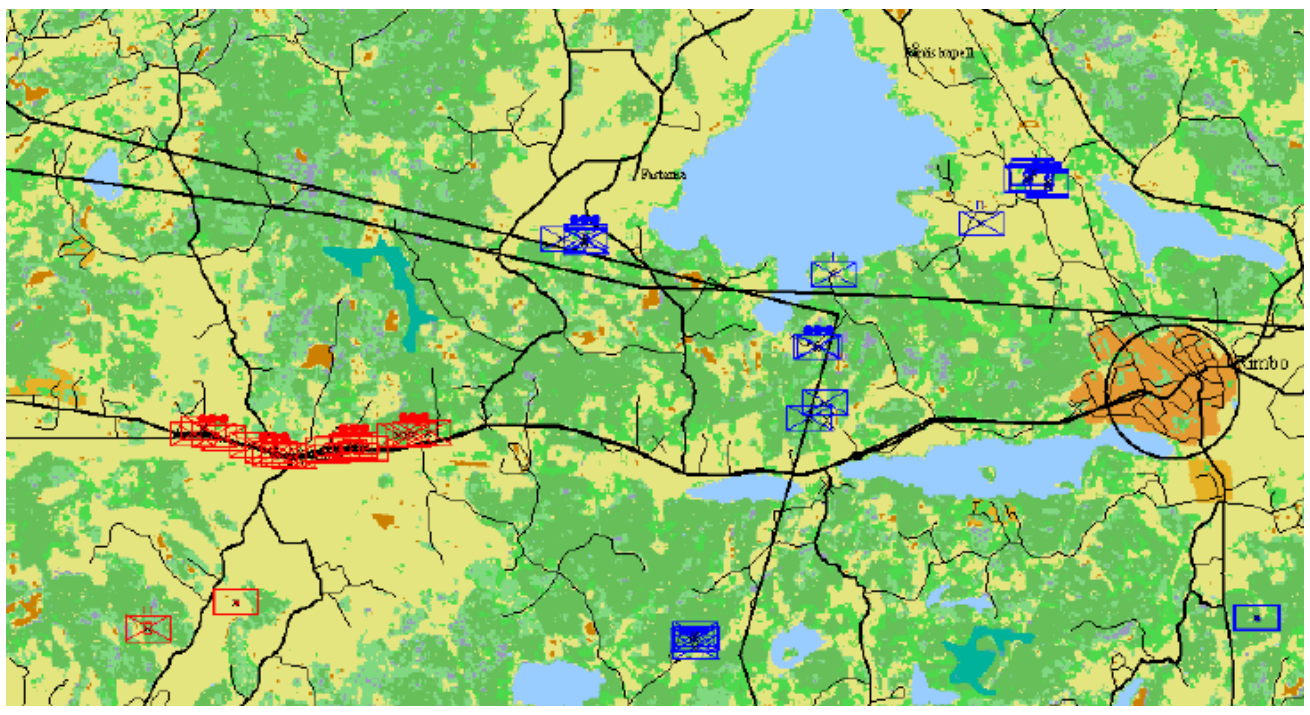

Figure 3: The *FbSim* simulator.

The simulated units, called actuators, are designed as agents, with decision and action capabilities. They have information (through sensors, communication with other agents, and a battle plan) and use a set of rules to make decisions based on this information, and then respond with weapons, vehicle actions or by giving orders. When a simulation is run, these agents act independently.

When a vehicle gets into a visual contact with another vehicle, a report is generated and saved in a log-file. Whether a visual contact is established depends on the distance between the two vehicles and the terrain between the two vehicles. The better the visual contact, the more specific classification of the vehicle is obtained.

### 2.3 *Intelligence reports*

In *FbSim*, when a vehicle gets a visual contact with another vehicle, a report is generated. This report is composed of the following slots: [from, name, position, time, classification, orientation]

- *from* is the name of the observer
- *name* is the name of the observed vehicle (only to be used for validation and not in the aggregation process)
- *position* is the position of the observed vehicle
- *time* is the number of seconds since simulation start
- *classification* is the type of the observed vehicle, at a long range the classification could be *unknown*, then perhaps *tracked vehicle* and finally a *tank*
- *orientation* is the direction of the observed vehicle.

### 2.4 *What do we want to know?*

An understanding of the current situation is essential for evaluating threats and courses of action. A reasonably accurate picture, abstracted from individual intelligence reports spread out in time, of the current number and positions of the opposing forces is itself valuable as it gives a measure of the scale of the threat and a guide to possible objectives.

When it comes to incapacitating opposing units with methods that require precision, single units or tightly spaced groups of units is perhaps the most useful level of analysis of a situation. After all weapon deployment seeks to

1. FbSim was jointly developed by the Defence Research Establishment, the Swedish Defence Materiel Administration, Bofors AB, Ericsson Microwave Systems AB, Celsius Aerotech AB, Saab Dynamics AB and Sjöland & Thyselius Datakonsulter AB.

incapacitate physical objects and while the ambition may be to incapacitate an organizationally more complex unit, this can be done only by incapacitating a sufficient number of the concrete physical objects that constitute the complex unit.

However if one seeks understanding of an ongoing development, the relevant level of analysis is typically not on the scale of single physical objects but on organizationally relevant groups of objects. Often enough the objectives with a military assignment are such that no particular single unit is necessary for the assignment to be carried through. Furthermore a group of single units can pose a qualitatively different threat–a far greater threat–than the sum of the individual threats posed by the single units. Thus quite apart from the need to present a situation to a decision maker in a way that avoids clutter, grouping vehicles together into units is an important step towards exposing the objectives of the opposing side: an important step in finding effective counter-measures.

Tracking individual physical objects on the ground can be extremely hard, in part due to irregularities in the terrain which reduces sensor coverage but often also in part due to the sheer number of very similar objects within a relatively small area.

When the opposing forces display organizational structure and when this structure is at least partially known one can go even further. To be able to classify a group of objects as being at least part of a unit of type *X* means that one can access previous information about units of type *X*: what function do units of that type fill in the opponents organization? what kind of capacity as regards movement and fire-power do they have? how is the unit itself organized and what kind of behavior can one expect from the unit? In addition, if one has only observed parts of the unit, a correct classification gives one reason to suspect that the remaining parts of the unit should be somewhere in the vicinity, which shows that the threat may be greater than it seems and gives one reason to direct sensor- and other resources towards the area.

## 3 Scenario

Our scenario consists of two forces, the red force is moving on a road towards a small town, the blue force is defending the town (see Figure 3). The area is about 10 000 x 10 000 meters, the terrain is partially hills and forests. Both forces consist of one company (see section 2.1). Both companies consist of three mechanized platoons, each with four armored personnel carriers (tracked). The blue company also contains a anti-tank robot platoon, with five anti-tank missile launchers. The red company also contains an MBT platoon, with five MBTs. As a consequence we have 17 vehicles on both sides.

Not all vehicles are detected and are hence not reported. The scenario takes about 13 min. and all together, 356 reports are generated, which gives an average of 27 reports/min. There are 14 blue vehicles detected that cause 204 reports and eight red vehicles detected that cause 152 reports. So at best our aggregation algorithm, would say that there are 14 blue vehicles and eight red vehicles. The first thing we can do is to plot all reports in IS Mark (see Figure 4).

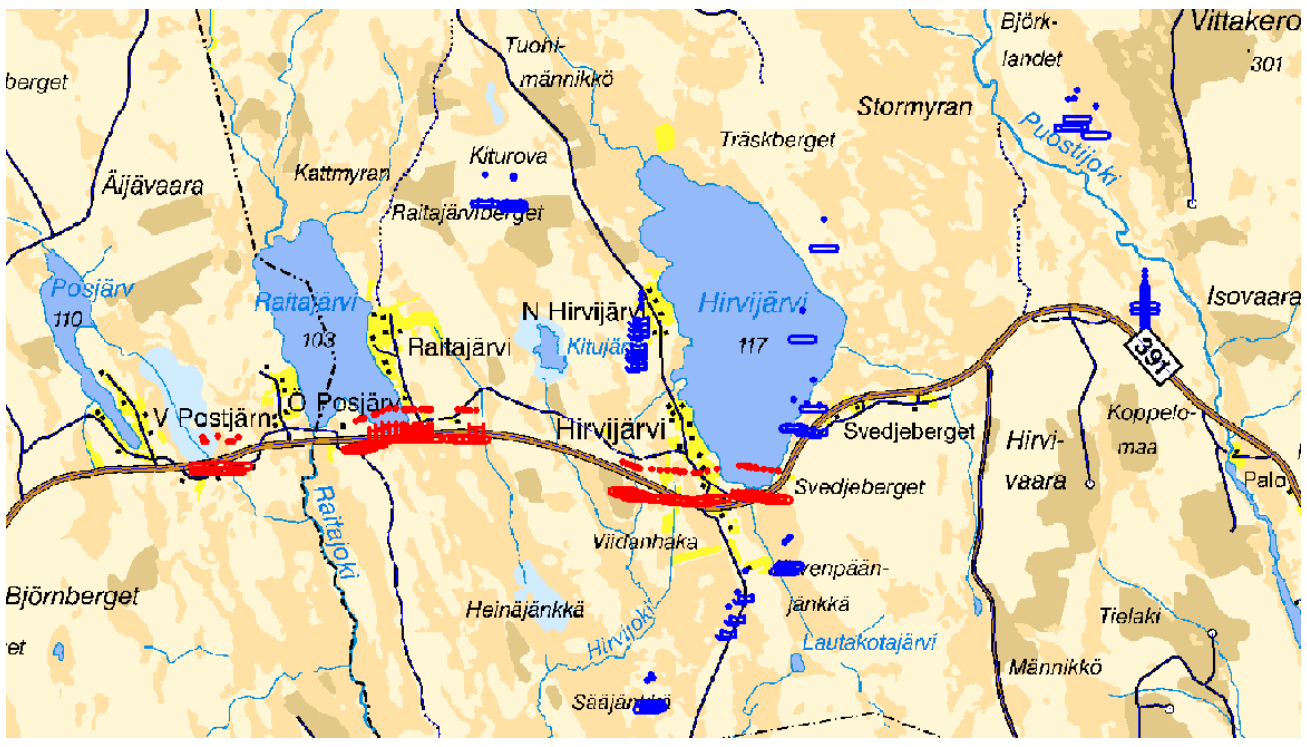

Figure 4: All reports in IS Mark.

## 4 Aggregation

Aggregation can take place on all levels: from intelligence reports to vehicles, from vehicles to platoons, and from any size of units to higher-up units in the military hierarchy. In this paper we focus only on the first three levels: intelligence reports, vehicles and platoons.

In aggregation it is just as crucial to combine the correct lower elements of all that are possible, as it is to combine those elements into a correct higher element. When there are many different lower elements available in comparison to the average number of elements in a template we face an initial selection problem of choosing which elements should be combined together for many different parallel aggregations, Figure 5.

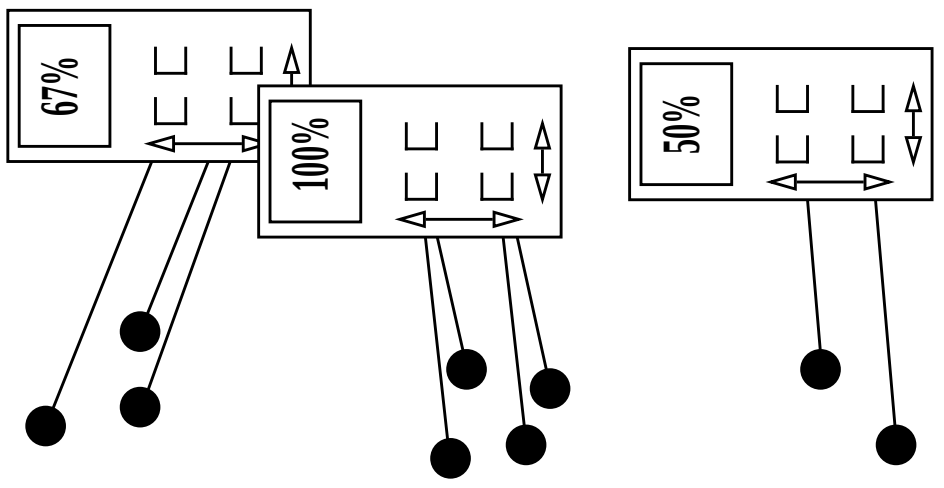

Figure 5: Parallel aggregation.

Secondly, when there are several different types of templates we must select which template to choose for these elements. Obviously, these two phases are not independent. Choosing which elements to fuse together depends very strongly on the template. We must avoid combining elements that do not match any of the templates.

If this is the case, why do we not use the templates from start to control which elements to fuse?

The answer is simple. While there is nothing to prevent us from doing that from a theoretical point of view it is unwise from a computational point of view. In order to partition a large set of elements into groups which should be fused we prefer to handle only pairwise relations between the elements. As the templates are usually relations between more that two elements their use would have a very high computational load.

Instead we use a pairwise distance measure between each pair of elements [5] that have some of the most important characteristics of the template. We define different conflict measures between intelligence reports and between vehicles, respectively. When any of these are violated a conflict arises between two elements, section 4.1. The different conflicts from all violated conditions are combined with Dempster's rule into an overall conflict for the two elements and serve as the actual distance measure for both clustering and classification.

If the conflict is high among all the elements of the template the hypothesis (corresponding to the template) must be disregarded. In both processes the criterion to find the best aggregation is to minimize the conflict. With this we first find which elements fit together by using a very fast clustering method described in 4.2. As the clustering is based on some of the most important characteristics of the templates on the next level the cluster result is usually quite good on average, but not necessarily perfect. Secondly, we classify those that do fit together by using templates on the clustering result, as described in section 4.3.

## 4.1 *Conflict*

The types of objects that can be aggregated are reports, vehicles and units, e.g., reports are aggregated into vehicles. Thus it is always assumed that a vehicle has caused the reports. Furthermore, vehicles can be clustered into units; i.e., the vehicles all belong to some unit. And finally, any size of units can be aggregated into larger units.

There are different aspects of conflicts, depending on the characteristics of the two objects. Aspects of a report conflict could be conflict with regard to type, direction or position. Instead of a position conflict we use a speed conflict by calculating the speed at which a vehicle must travel, in order to cause the two reports.

### 4.1.1 *Reports*

A physical conflict between two reports is the *speed conflict*. If a vehicle must travel at a greater speed than is physically (type of vehicle, terrain) possible, in order to cause the reports, there is a conflict between the reports. Another physical conflict is the *type conflict*. There is a conflict between two reports if they describe two different types of vehicles, but there is no conflict if one report indicates an unknown type. The third type of conflict is the *direction conflict*. If two reports describe two vehicles, traveling in opposite directions, there is a conflict between the reports, and no conflict if they travel in the same direction.

The conflict is a value between 0 and 1, where 0 means no conflict and 1 means an absolute conflict. The overall conflict is obtained by combining the conflicts of the different aspects according to

$$C = 1 - \prod_{\forall a}(1 - C_a). \tag{1}$$

### *Speed Conflict*

The speed conflict is given by calculating the speed a vehicle must have in order to cause the two reports. The speed conflict is then given by eq. (2) and shown in Figure 6. We have

$$conf(x, p, x_1, x_2) = $$

$$= \begin{cases} \dfrac{xp}{x_1}, & x < x_1 \\ \dfrac{(x - x_1) + p(x_2 - x)}{x_2 - x_1}, & x_1 \le x < x_2 \\ 1, & x > x_2 \end{cases}. \tag{2}$$

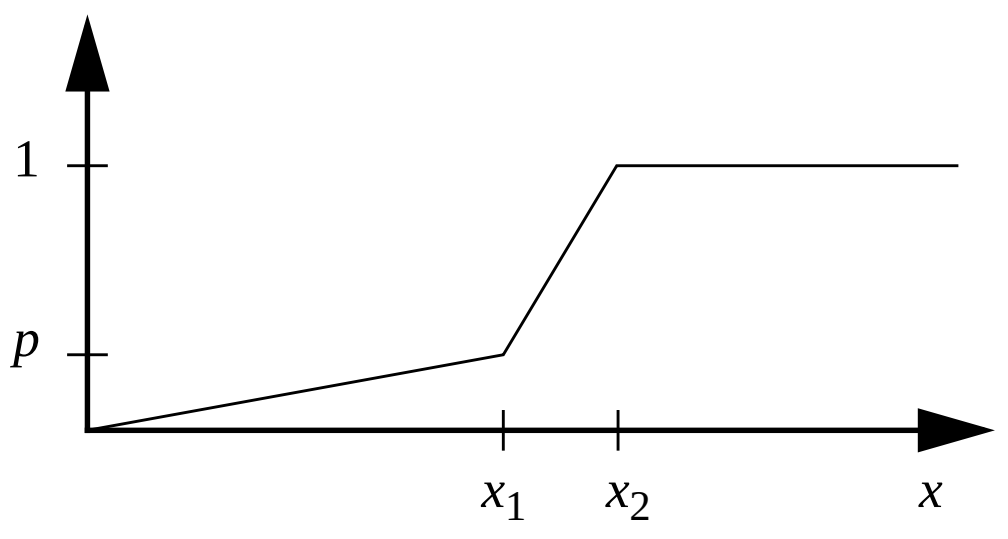

Figure 6: The conflict function.

The parameter $x$ is given the speed value. If we knew the maximum speed at which a particular vehicle could travel, given a certain terrain, we could set the conflict to one for values greater than the maximum speed. However, this is not the case. What we can say is that $x > x_2$ is impossible and $x < x_1$ is certainly possible. We would also like to express that it is more probable that two reports belong to the same vehicle, the lower the speed requirement is. Therefore we introduce the parameter $p$. We used the following values: $p$ = 0.01, $x_1$ = 22 m/s, $x_2$ = 25 m/s.

### *Type Conflict*

All vehicle classifications are hierarchically ordered in a tree, with *unknown* as root and the specific vehicle types as the leaves. The conflict function then returns zero if one classification is a descendant of the other and otherwise returns one. We used the hierarchy shown in Figure 7.

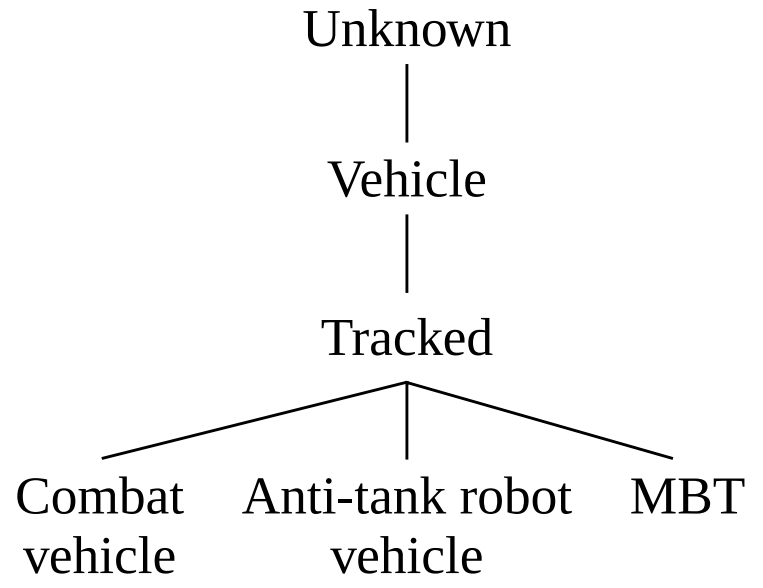

Figure 7: Classification of vehicles.

### *Direction Conflict*

The direction conflict is not an absolute conflict, since it is possible for a vehicle to change direction between two reports. Assume we have ten reports describing ten identical vehicles that are close to each other. Five reports describing a vehicle moving south and five reports describe a vehicle moving north. It is then more likely that two vehicles have caused the reports instead of one.

Minor variations in the directions should be allowed, since the vehicles, for instance, might be following a non-straight road. Also, if too long time has elapsed between the two reports, the direction conflict becomes obsolete, eq. (3),

$$conf_\delta(\delta d, \delta t, \delta d_0, \delta t_0, k) = $$

$$= \begin{cases} \dfrac{k\ \delta d}{\pi(k + \delta t)}, & (\delta t \le \delta t_0) \wedge (\delta d \ge \delta d_0) \\ 0, & \text{otherwise} \end{cases} \tag{3}$$

where $\delta d$ is the difference in directions and $\delta t$ is the difference in time. If $\delta d < \delta d_0$ the difference in direction is considered to be to small, to influence the conflict. If $\delta t > \delta t_0$ to long time has elapsed, to influence the conflict.

We use the following values: $\delta d_0 = \pi / 4$ and $\delta t_0 = 8\,\mathrm{s}$. The parameter $k$ is used to get a slower increase of the conflict. We use the value $k = 10$.

### 4.1.2 *Vehicles*

When vehicles are to be clustered into units, there are no physical conflicts since the structure of a unit depends on templates. For instance a tank platoon template contain five tanks, driving 50–200 meters apart. All vehicles in a unit also tend to drive in the same direction. So, when clustering vehicles into units, we consider their relative distance and their direction.

A difficulty is that we do not know where the vehicles are, for every specific time. We only know where they have been, according to reports. The same holds for their direction. Instead we plot the vehicle position according to its reports and combine them with lines, as shown in Figure 8.

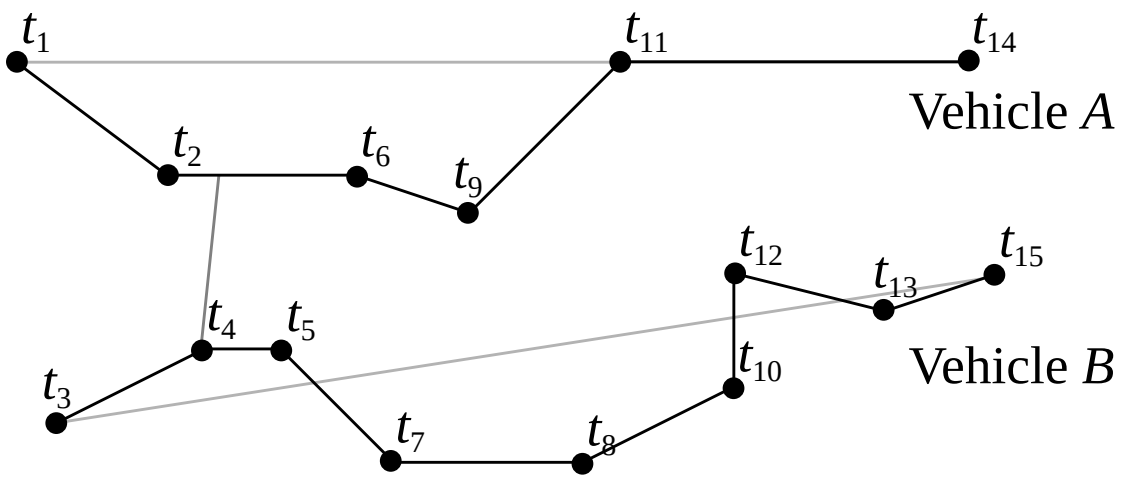

Figure 8: Plotting the reports.

In Figure 8 we have two vehicles with a total of 16 reports, with time stamps $t_1$ to $t_{15}$, where $t_i < t_j$ for $i < j$.

#### *Distance Conflict*

There are a number of different distances that could be calculated. The most trivial would be to calculate the distance between the respective vehicles last known positions (last reports positions). However, a better approach is to consider all reports. We know, for instance, where vehicle $B$ is at $t_4$. We do not know where vehicle $A$ is at $t_4$, but we do know where vehicle $A$ is at $t_2$ and $t_6$. If we assume that vehicle $A$ travels in a straight line and at constant speed, we can calculate where vehicle $A$ is at $t_4$. We use this method to calculate all distances at time $t_3$ to $t_{14}$, the maximum starting time and minimum finishing time, respectively, of the two vehicles and get twelve distances. We call the time interval $t_3$ to $t_{14}$ the common time interval. A distance candidate to use in the conflict function could be the minimum distance, the maximum distance, the average distance or the median distance. We used the median distance to avoid that a few wrongly clustered reports would influence the result.

The distance is then fed into the conflict function given by eq. (2). Here, we used: $p = 0.01$, $x_1 = 300$ m, $x_2 = 1000$ m, with $x$ the median distance.

#### *Direction Conflict*

To calculate the difference in direction for vehicle $A$ and vehicle $B$, based on their respective reports, we calculate the direction vehicle $A$ must take in order to travel at a straight line from its position at $t_3$ to its position at $t_{14}$. We do the same for vehicle $B$ and calculate the difference in direction and feed the value into eq. (2). We use the following values: $p = 0$, $x_1 = 0$, $x_2 = \pi$, with $x$ the direction.

## 4.2 *Clustering*

The clustering method developed may cluster intelligence reports, vehicles and units on all hierarchical levels. Initially, we consider intelligence reports regarding observations of vehicles that come from multiple sources. Here, it is not known apriori if two different intelligence reports refer to the same vehicle. All reports concerning one vehicle should be fused separately from all other intelligence reports.

We use the clustering process to separate the intelligence into subsets for each vehicle. We combine Dempster-Shafer theory with the Potts Spin Neural Network model [17] into a powerful solver for very large Dempster-Shafer clustering problems [1]. The Potts model has proven useful in many complex optimization problems [4]. We believe this method can serve as a general solution for preprocessing of intelligence data in information fusion.

Let $i$ be an index for the element and $a$ the index for the clusters. Then $S_{ia} = 0, 1$ is a discrete vector with the constraint $\sum_{a=1}^{q} S_{ia} = 1\ \forall i$ where $S_{ia} = 1$ means that element $i$ is in cluster $a$. Then the energy function that defines the Potts model is

$$E = \frac{1}{2} \sum_{i,j=1}^{N} \sum_{a=1}^{q} J_{ij} S_{ia} S_{ja} . \qquad (4)$$

This model can serve as a data clustering algorithm if $J_{ij}$ is used as a penalty factor of element $i$ and $j$ being in the same cluster; elements in different clusters get no penalty.

The problem consists of minimizing this energy function by changing the states of the $S_{ia}$. This process is carried out with simulated annealing. Simulated annealing uses temperature as an important factor. We start at a high temperature where the $S_{ia}$ change state more or less at random, and are only marginally biased by their interactions $J_{ij}$. As the temperature is lowered parts of the system become constrained in one way or the other, they freeze. Finally, when the complete system is frozen, the spins are completely biased by the interactions ($J_{ij}$) so that, hopefully, the minimum of the energy function is reached. For computational reasons we will use a mean field model, where spins are deterministic [4].

In each cluster we use the conflict of Dempster's rule when all elements within a subset are combined as an indication of whether these elements belong together. The higher this conflict is, the less credible that they belong together.

In [5] a criterion function of overall conflict called the metaconflict function was established. The metaconflict was derived as the plausibility that the partitioning is correct.

Definition. *Let the* metaconflict function,

$$Mcf(q, S_1, S_2, \ldots, S_n) \triangleq 1 - \prod_{i=1}^{q} (1 - c_i), \qquad (5)$$

*be the conflict against a partitioning of n pieces of evidence of a set* $\chi$ *into q disjoint subsets* $\chi_i$*. Here,* $c_i$ *is the conflict in subset i.*

In Dempster-Shafer clustering we use the minimizing of the metaconflict function as the method of partitioning all elements into separate subsets. After this, each subset refers to a different event and the reasoning can take place with each event treated separately.

We will logarithmize the metaconflict in order to be able to use it as interaction in the Potts model.

We have

$$\begin{aligned} &\min Mcf \\ &\Leftrightarrow \\ &\max \log(1 - Mcf) = \max \log \prod_i (1 - c_i) \qquad (6) \\ &= \max \sum_i \log(1 - c_i) = \min \sum_i -\log(1 - c_i) \end{aligned}$$

where $-\log(1 - c_i) \in [0, \infty]$ is a weight [16] of evidence, i.e., in this context a weight of conflict.

Since the minimum of Mcf (= 0) is obtained when the final sum is minimal (= 0), the minimization of the final sum yields the same result as a minimization of Mcf would have done.

In Dempster-Shafer theory one defines a simple support function, where the evidence points precisely and unambiguously to a single nonempty subset $A$ of $\Theta$. If $S$ is a simple support function focused on $A$, then the basic probability numbers are denoted $m(A) = s$, and $m(\Theta) = 1 - s$. If two simple support functions, $S_1$ and $S_2$, focused on $A_1$ and $A_2$ respectively, are combined, the weight of conflict between them is

$$\mathrm{Con}(S_1, S_2) = \begin{cases} -\log(1 - s_1 s_2), & A_1 \cap A_2 = \varnothing \\ 0, & \text{otherwise} \end{cases} \qquad (7)$$

which may be rewritten as

$$\mathrm{Con}(S_1, S_2) = -\log(1 - s_1 s_2)\,\delta_{|A_1 \cap A_2|} \qquad (8)$$

with $\delta_{|A_1 \cap A_2|}$ being defined so that it is equal to one for $A_1 \cap A_2 = \varnothing$ and zero otherwise.

The complete energy function including constraints that we are considering is

$$E[S] = \frac{1}{2}\sum_{a=1}^{K}\sum_{i,j=1}^{N} J_{ij}S_{ia}S_{ja} - \frac{\gamma}{2}\sum_{a=1}^{K}\sum_{i=1}^{N} S_{ia}^2 + \frac{\alpha}{2}\sum_{a=1}^{K}\left(\sum_{i=1}^{N} S_{ia}\right)^2 \quad (9)$$

where the first term is the standard clustering cost. Using mean field theory we can find the minimum of this energy function. The Potts mean field equations are derived as [4]:

$$V_{ia} = \frac{e^{-H_{ia}[V]/T}}{\sum_{b=1}^{K} e^{-H_{ib}[V]/T}} \quad (10)$$

where

$$H_{ia}[V] = \frac{\partial E[V]}{\partial V_{ia}} = \frac{\sum_{j=1}^{N} J_{ij}V_{ja} - \gamma V_{ia} + \alpha \sum_{j=1}^{N} V_{ja}}{G_a} \quad (11)$$

with $V_{ia} = \langle S_{ia} \rangle$.

In order to minimize eq. (9) we use eqs. (10) and (11) recursively until a stationary equilibrium state has been reached for each temperature. To apply it to Dempster-Shafer clustering we use interactions $J_{ij} = -\log(1 - s_i s_j)\,\delta_{|Ai \cap Aj|}$.

The algorithm for simulating these spins works roughly as follows: Use a precomputed highest critical temperature, $T_c$, as the starting temperature. Choose the mean field spins to be in their symmetric high temperature state; $V_{ia} = 1/K\ \forall i, a$. At each temperature, iterate eqs. (10), (11) until a fix point has been reached. The temperature is lowered by a constant factor until every spin has frozen, i.e., $V_{ia} = 0, 1$, Figure 9.

In order to find the correct number of clusters the parameter $K$ must be varied and the remaining conflict after clustering evaluated against some threshold. When $K$ is increased the remaining conflict after clustering decreases. When it falls below the threshold the best $K$ is considered found. In Figure 10 an example of clustering 204 intelligence reports is investigated. Here, the logarithm of the total weight of conflict for 1–20 clusters is calculated. The total weight of conflict falls below the threshold for $K = 10$.

INITIALIZE

$K$ (# clusters); $N$ (# elements);

$J_{ij} = -\log(1 - s_i s_j)\,\delta_{|A_i \cap A_j|} \quad \forall i, j$;

$s = 0$; $t = 0$; $\varepsilon = 0.001$; $\tau = 0.9$; $\alpha$ (for $K \le 7$: $\alpha = 0$, $K = 8$: $\alpha = 10^{-6}$, $K = 9$: $\alpha = 0$, $K = 10$: $\alpha = 3 \cdot 10^{-7}$, $K = 11$: $\alpha = 3 \cdot 10^{-8}$); $\gamma = 0.5$;

$T^0 = T_c$ (a critical temperature) $= \frac{1}{K}max(-\lambda_{min}, \lambda_{max})$, where $\lambda_{min}$ and $\lambda_{max}$ are the extreme eigenvalues of M, where $M_{ij} = J_{ij} + \alpha - \gamma\delta_{ij}$;

$V_{ia}^0 = \frac{1}{K} + \varepsilon \cdot rand[0,1] \quad \forall i, a$;

REPEAT

- REPEAT–2
  - $\forall a\ G_a^s = \frac{K}{N} \cdot \sum_{i=1}^{N} V_{ia}^s$;
  - $\forall i$ Do:
    - $H_{ia}^s = \frac{\sum_{j=1}^{N}(J_{ij} + \alpha)V_{ja}^s - \gamma V_{ia}^s}{G_a^s} \forall a$;
    - $F_i^s = \sum_{a=1}^{K} e^{-H_{ia}^s/T^t}$;
    - $V_{ia}^{s+1} = \frac{e^{-H_{ia}^s/T^t}}{F_i^s} + \varepsilon \cdot rand[0,1] \forall a$;
  - $s = s + 1$;

  UNTIL–2

  $\frac{1}{N}\sum_{i,a}\left|V_{ia}^s - V_{ia}^{s-1}\right| \le 0.01$;
- $T^{t+1} = \tau \cdot T^t$;
- $t = t + 1$;

UNTIL

$\frac{1}{N}\sum_{i,a}(V_{ia}^s)^2 \ge 0.99$;

RETURN

$\{\chi_a \mid \forall S_i \in \chi_a.\ \forall b \ne a\ V_{ia}^s > V_{ib}^s\}$;

Figure 9: The clustering algorithm.

*correction inserted: figure 9*

In Figure 11 and 12 the clustering process of clustering 204 intelligence reports into ten clusters is illustrated. In Figure 11 the neuron output from 2040 neurons over 19 iterations is shown. Each intelligence report is represented by ten neurons showing the degree to which the intelligence report belongs to the clusters. As shown, most of the clustering takes place after the twelfth iteration. In Figure 12 the seven last iterations are shown. Each rectangle corresponds to one iteration, the ten columns correspond to ten different clusters and each row corresponds to one report. The output signal of a neuron is indicated by the size of the square. In the final iteration (rightmost rectangle) all values are close to one and the clustering has terminated. In the top row of the rightmost rectangle we find that intelligence report number one is in cluster number two, etc.

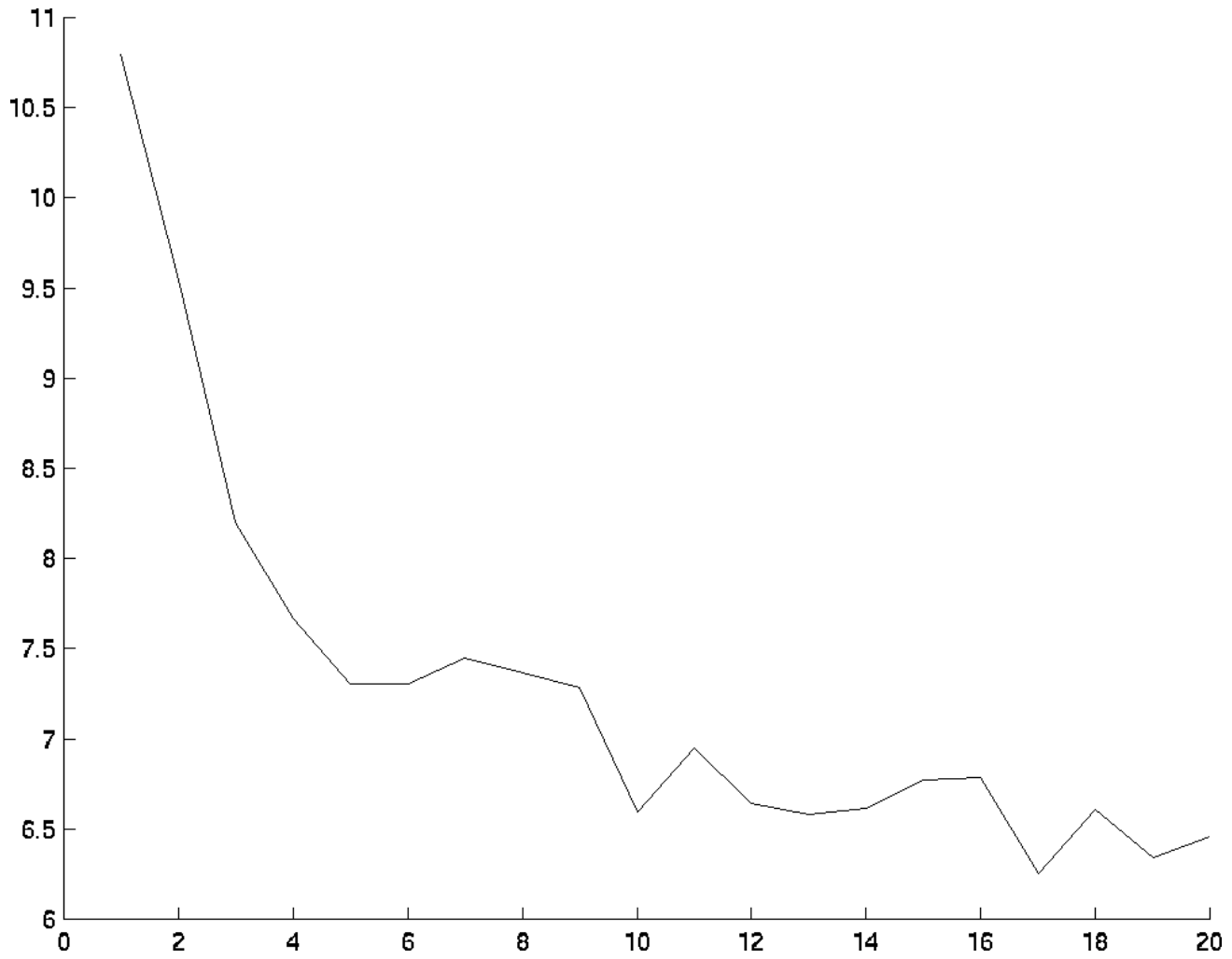

Figure 10: The logarithm of total weight of conflict when clustering into 1–20 clusters.

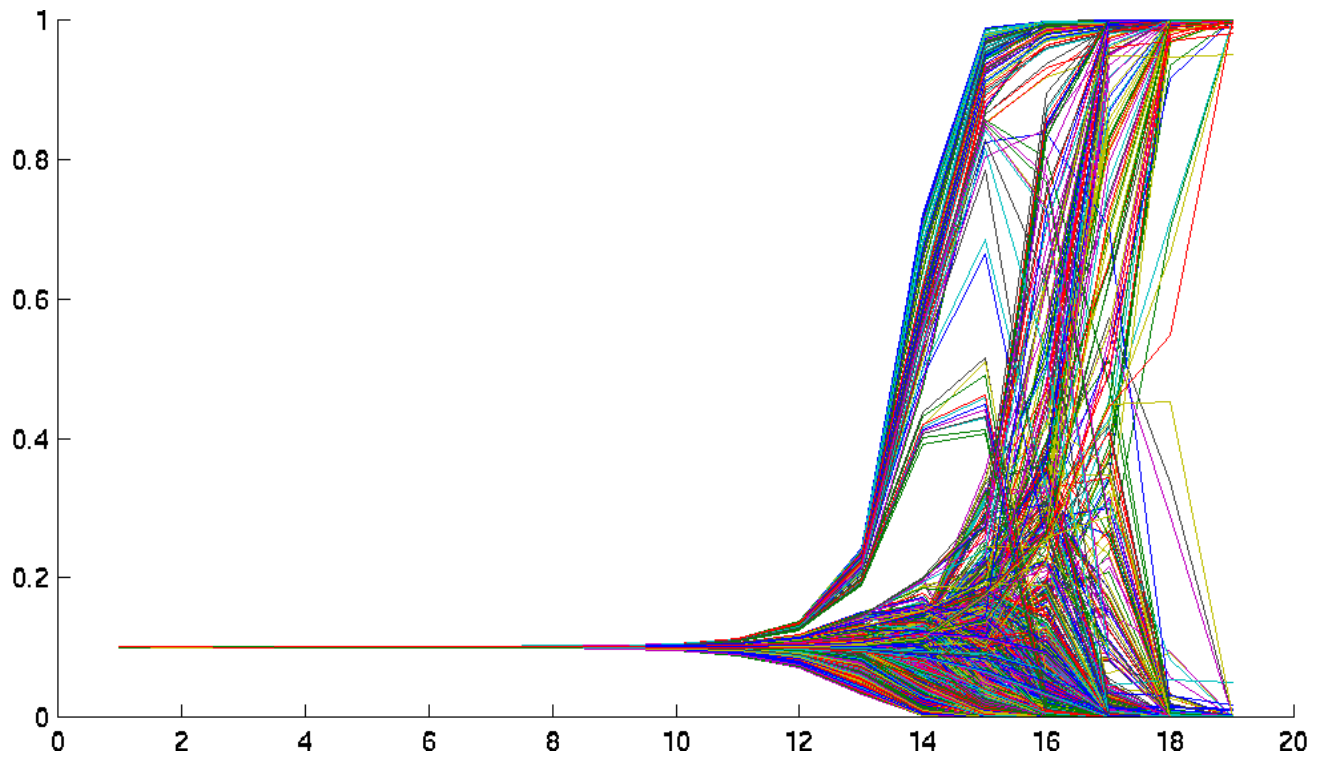

Figure 11: Neuron output from 2040 neurons over 19 iterations.

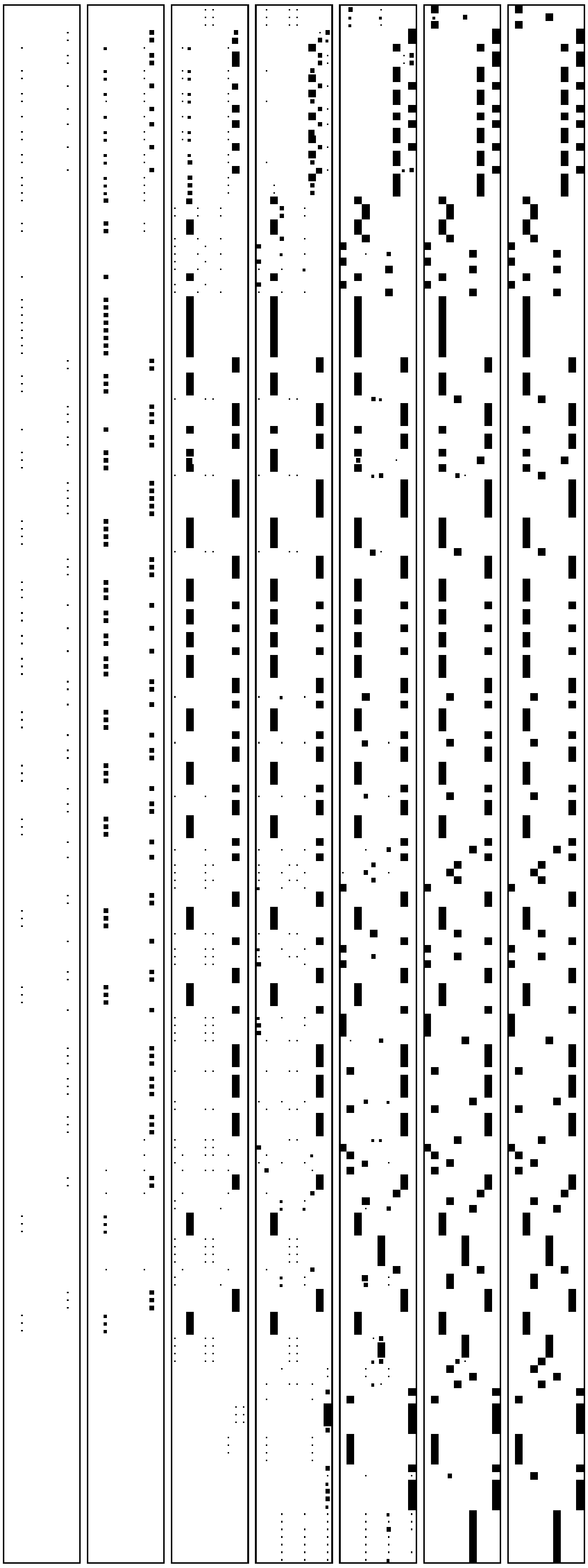

Figure 12: From left to right: The seven last iterations of the clustering process.

On a test problem when clustering $N$ (= $2^K - 1$) pieces of evidence into $K$ subsets where the evidence supports all subsets of the frame $\Theta = \{1, 2, 3, \ldots, K\}$ the Potts spin Dempster-Shafer clustering was shown to have a computational complexity of $O(N^2 \log^2 N)$.

A further description of the cluster methods used in intelligence analysis was given in [15].

## 4.3 *Classification*

The next goal is to combine vehicles into units and classify the units. While we employ some basic methods and terminology from the Dempster-Shafer framework the methods involved are quite different from the methods described in the previous section.

Units are described in templates. A unit template might state that a unit of type $X$ consists of three vehicles of type $A$ and two vehicles of type $B$.

*Hypotheses* are of the form "vehicles $v_1, \ldots, v_n$ together form a unit of type $T_1$ or ... or $T_m$". In such an hypothesis it is implicitly understood that no other observed vehicle than $v_1, \ldots, v_n$ belongs to the unit (but of course there may be other units of the same type present). Hypotheses may *conflict*, this happens if and only if a vehicle occurring in one hypothesis also occurs in the other.

The overall objective is to find the best consistent (no hypotheses conflict) and complete (every vehicle is accounted for) set of hypotheses. This involves generating and assessing the worth of not only single hypotheses but sets of hypotheses.

Note that an hypothesis is disjunctive in character so while a number of hypotheses regarding which objects belong to the same unit will be discarded along the way no commitment is made to the exact unit-type classification of a group of objects until it is presented to the user. Even then judgement may be withheld if the competing alternative classifications are close enough: a disjunction can be presented to the operator.

### 4.3.1 *Generating hypotheses*

Hypotheses are generated as follows, Figure 13.

```
Step 1:Generate all hypotheses of size
   1.
Step 2:For each surviving hypothesis of
   size n and for each vehicle not in the
   hypothesis, generate a hypothesis of
   size n+1 by adding the vehicle to the
   hypothesis. Evaluate the hypothesis,
   if it is good enough, keep it.
Step 3:Repeat step 2 until no new
   hypotheses are generated.
```

Figure 13: Generation of hypotheses.

In the worst case the above algorithm will generate $2^n$ hypotheses. This is worrisome, in particular as all subsequent processing depends on the number of hypotheses generated. However, in the sample applications we have studied (20–100 vehicles) this has not been a problem (total run-time has been 2–10 seconds). There are mainly two reasons:

1) A careful pruning of the hypotheses generated–hypotheses involving vehicles that are too far apart are discarded, as are hypotheses where a vehicle of a particular type does not belong to a unit of a particular type (see Step 2 above).
2) The problem space can often be partitioned into a set of independent problem spaces that can be solved one by one (see below).

In the sample applications these factors have been sufficient to reduce the problem space to manageable proportions. In the general case, however, one can set an upper limit to the number of hypotheses of a particular size involving any particular vehicle. Taking the $m$ best hypotheses of any size for any vehicle means that at most $(m - 1)\, n^2$ hypotheses of any size will have to be generated for comparison. In this way the generating algorithm becomes polynomial in time.

### 4.3.2 *Evaluating hypotheses*

Hypotheses are evaluated with regard to two parameters:

1) How the type of vehicles and the number of vehicles of each type fits the description of a unit of that type.

2) How the positions of the vehicles fit the hypothesis that they belong to the same unit.

These are combined using the method of orthogonal combination from the Dempster-Shafer framework. An hypothesis is “good enough” if its conflict value is below some threshold.

In a disjunctive hypothesis each disjunct is evaluated independently of the others and the entire hypothesis is given the conflict value of the disjunct with the minimal conflict. This can be justified by appealing to the non-probabilistic nature of the conflict values: the values should be interpreted as degrees of fit. So for instance two disjuncts can both be given a value of say 0.75.

Evaluation occurs at two places: when hypotheses are generated one needs to estimate their worth and when sets of hypotheses are compared one needs to evaluate the worth of a set of hypotheses. These problems, of course, are not independent.

The general method described in the previous section leaves it open how evaluation is done, but it does impose some constraints. The computational need to prune off “bad” hypotheses means that it is not sufficient that we can rank hypotheses according to their worth. We need some criterion of *good* and *bad*, for instance, if numerical methods are employed, a threshold. For this reason we choose to rely on the Dempster-Shafer method of orthogonal combination rather than Bayesian conditionalisation. Below this will be given further motivation.

We should emphasize that the problem of hypothesis evaluation is still in need of development. One might say that at the present stage we are more interested in localizing and isolating the issues and problems involved than in finding completely satisfactory solutions to the problems.

A simple hypothesis (an hypothesis without disjunctions) makes two assertions: (1) the vehicles in the hypothesis satisfy the spatial constraints of being that part of a unit that has been observed, (2) the unit is of type *X*.

When we speak of the conflict of an hypothesis we refer to the degree to which the vehicles mentioned in the hypothesis deviate from an ideal unit of type *X*. Thus, we assume, in the ideal unit no vehicles are missing and they group themselves in a particular way. The more the vehicles deviate from this the higher the conflict assigned to the hypothesis.

In our present modelling we treat the formation/spacing conflict as independent of the classification conflict. Thus we do not take into consideration the possibility that different units may place different spatial constraints on the vehicles in the unit. This however should not be too difficult to correct.

### *Classification*

We have used a simple, indeed simplistic, method for classifying units. The fit is based on the ratio of observed vehicles/expected vehicles. Thus if the hypothesis is of the form “tanks *a*, *b* and *c* form a unit of type *X*” where a unit of type *X* in its standard setup consists of *four* tanks, the classification conflict is 0.25 (1 – ratio) and the support is 0.75. If the hypothesis is of the more indeterminate form “vehicle *a* and tanks *b* and *c* form a unit of type *X*” the classification conflict is still 0.25, but the direct support is only 0.5.

The problems with such a method are fairly obvious. No consideration is taken to base frequencies of unit types and no consideration is taken to the probability of detection or clutter. While it would be possible to let these factors influence the result this would have to be done in an *ad hoc* fashion as there is no underlying theory.

In several respects the present classification method is inferior to the Bayesian method proposed in [3] where base frequencies and probability of detection is dealt with in a principled manner. However, there are two problems with the Bayesian approach.

The first is how to establish the probability of detection. Quite apart from the fact that the probability of detection varies with the kind of sensors employed, the kind of objects that are to be detected, whether it’s day or night, the weather and so on, we have the problem that sensors move around and terrain, hence visibility, varies. To deal with the varying probability of detection across time one needs recourse to geographical information, detailed information about sensors, information about which routes are possible for which vehicles and more still. Even though these problems need not be insurmountable, they are problems that we have not addressed here.

A more fundamental problem can be described as follows. While a Bayesian method of force aggregation such as [3] can give us the probability of the data given the hypothesis *P(D|H)* it will not as easily give the probability of the hypothesis given the data *P(H|D)* which is what we are interested in (the data here is of the form "vehicles *a*, *b* and *c* have been observed", while the hypothesis is that these form a unit of some particular type). From Bayes' rule

$$P(H|D) = \frac{P(H)P(D|H)}{P(D)} \quad (12)$$

we need (1) the prior probability of the data *P(D)*, and (2) the prior probability of the hypothesis *P(H)*. Estimating *P(D)* seems hopelessly difficult, but if we know the relative frequency of different unit types and if we were only interested in comparing *P(H|D)* for different hypotheses *H*, then *P(D)* merely becomes a normalizing constant.

Now the vehicles mentioned in *D* is a subset of the total number of vehicles observed. It is an hypothesis in its own right that the vehicles in *D* together form a unit. There might be a different method of carving out units which gives us a different set of data *D´* (for instance "vehicles *a*, *b* and *d* have been observed") which means that we must be able to compare the different ways of carving out objects which in turn means that we cannot treat *P(D)* as a normalizing constant. Remember: we are using the results from classification to determine an optimal way of clustering the data. The latter problem is not addressed in [3].

There is a further possibility for the Bayesian approach. If one can estimate the probability that the data is a result of a hitherto unknown unit type one can determine the value of *P(H|D)* (as then one has an exhaustive list of hypotheses). However without recourse to any deeper analysis of the probability of an unknown unit type one ends up applying a method with a firm theoretical basis to assumptions that are more or less arbitrary. It is not obvious that anything has been gained by such a move.

***Combining classification and formation conflict***

The formation conflict of a group of vehicles is treated as the average pairwise conflict of the vehicles (as described in section 4.2). Again there are other conceivable methods but this particular method does not penalize larger units which a number of other methods of combination do.

The classification conflict $c_0$ and the formation conflict $c_1$ using the probabilistic sum giving the hypothesis conflict

$$C(H) = 1 - (1 - c_0)(1 - c_1). \quad (13)$$

### 4.3.3 *Partitioning the problem space*

Two hypotheses *A* and *B* belong to the same *sub-problem* if there are hypotheses $H_0$, ..., $H_n$ such that *A* and $H_0$ are jointly inconsistent (conflict), $H_i$ and $H_{i+1}$ are jointly inconsistent for each $i < n$, and $H_n$ is inconsistent with *B*. Thus *A* and *B* need not be jointly inconsistent to belong to the same sub-problem, it is sufficient that there is a chain of inconsistency linking them.

The algorithm used for partitioning the problem space is fairly obvious: take any hypothesis and place it and all the hypotheses that are inconsistent with it in a partition, repeat the process for each new hypothesis added to the partition until no new hypotheses are added. The result is a partition. Take one of the remaining hypotheses and construct a new partition using the same method and repeat until all hypotheses have been added to some partition, Figure 14. The complexity of the algorithm is $O(n^2)$ where *n* is the number of hypotheses.

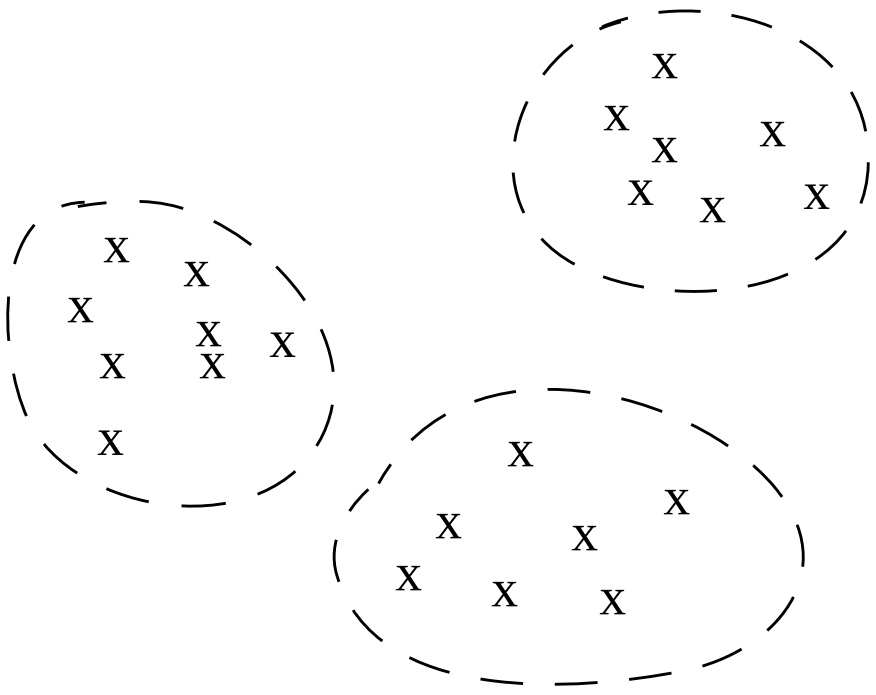

Figure 14: A problem space that has been divided into three sub-problems. Each x represents a vehicle.

#### 4.3.4 *Generating maximal consistent sets*

For each sub-problem one generates the set of maximal and consistent sets of hypotheses and selects the best such set. The complexity of the problem is reduced by the fact that to check whether a set of hypotheses is consistent it is enough to check all pairs of hypotheses are jointly consistent (this is due to the particular nature of the hypotheses).

The algorithm employed can be described as a recursive procedure F, Figure 15, that takes three lists of hypotheses and returns an optimal, maximal and consistent list of hypotheses. The initial call to F is F(nil, $S$, nil), where $S$ is a partition of the problem space.

```
F(CURRENT, REST, BEST):

IF for every object there is an hypothesis in CURRENT
    where the object occurs
THEN RETURN CURRENT or BEST depending on which
    has lowest conflict
ELSE
    FOR each H_i in REST
        CURRENT(H_i) : = APPEND(CURRENT, H_i);
        REST(H_i) : = {H_j | H_j is consistent with
                        CURRENT(H_i) and i < j};
        BEST : = F(CURRENT(H_i), REST(H_i), BEST);
    RETURN BEST
```

Figure 15: Procedure F.

The conflict value of a set or list of hypotheses $S = \{H_0, ..., H_m\}$ is given by eq. (14)

$$C(S) = 1 - [1 - C(H_0)] \cdots [1 - C(H_m)] . \quad (14)$$

The above algorithm has an upper bound of complexity $O(kn^4)$ where $n$ is the number of hypothesis and $k$ some constant.

## 5 Results

### 5.1 *Quantitative analysis*

First, let us investigate aggregation of intelligence reports into vehicles. After running our aggregation algorithm we end up with the following. The blue side: five armored personnel carriers (tracked), three unspecified tracked vehicles and two anti-tank robot vehicles. The red side: four armored personnel carriers (tracked), one unspecified tracked vehicle and four MBT (see Figure 16).

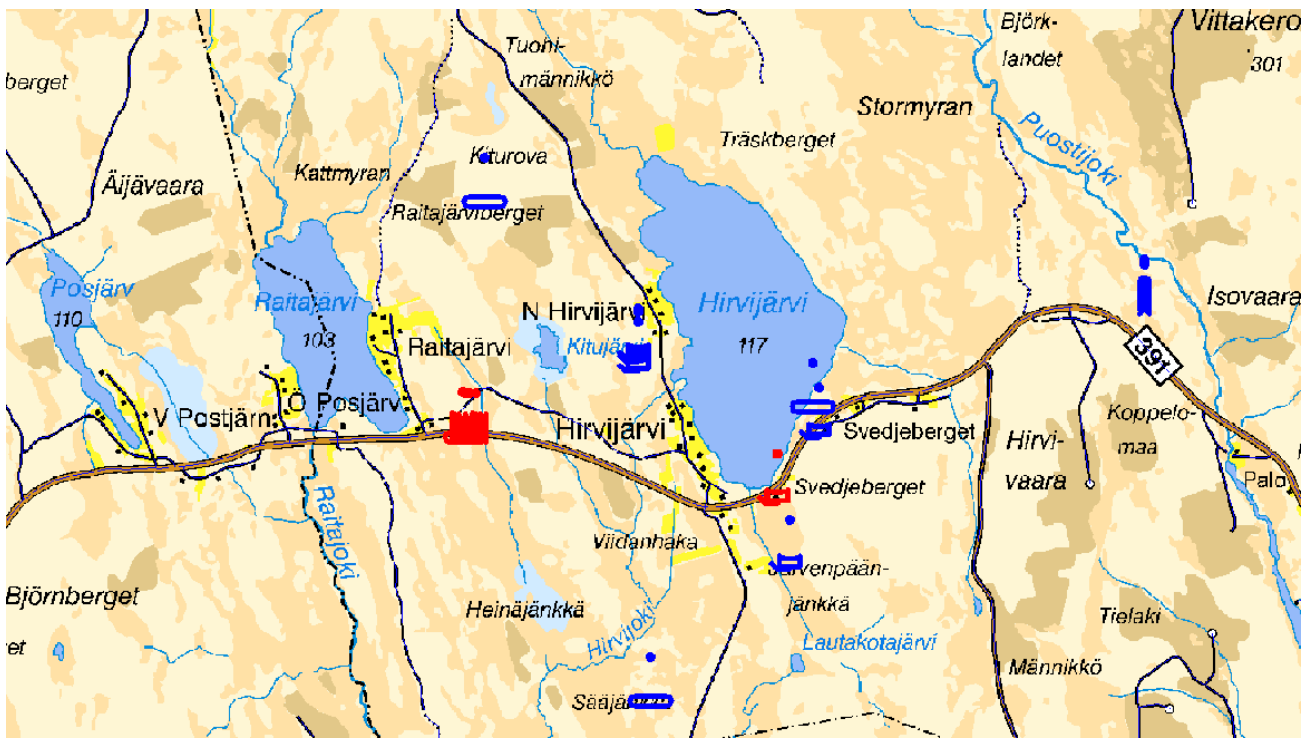

Figure 16: All vehicles in IS Mark.

Hence the aggregation algorithm concludes that there are ten blue vehicles, when there are actually 14 vehicles within sight (of the 17 in the scenario). This is due to the fact that the algorithm is based on minimizing conflicts between reports and that ten vehicles are sufficient to cause all blue reports. The algorithm also concludes nine red vehicles, when there are actually eight within sight. Here eight vehicles could have explained the reports, but that was deemed unlikely (too close to conflict).

The next step is to aggregate the vehicles into platoons. On the blue side, three armored personnel carriers (tracked) and one unspecified tracked vehicle are aggregated into a mechanized platoon, the two anti-tank robot vehicles and the two remaining unspecified tracked vehicles are aggregated into a anti-tank robot platoon, while the final two armored personnel carriers (tracked) remain unaggregated. On the red side, the four MBTs and one unspecified tracked vehicle are aggregated into an MBT Platoon and the four armored personnel carriers (tracked) are aggregated into a mechanized platoon (see Figure 17).

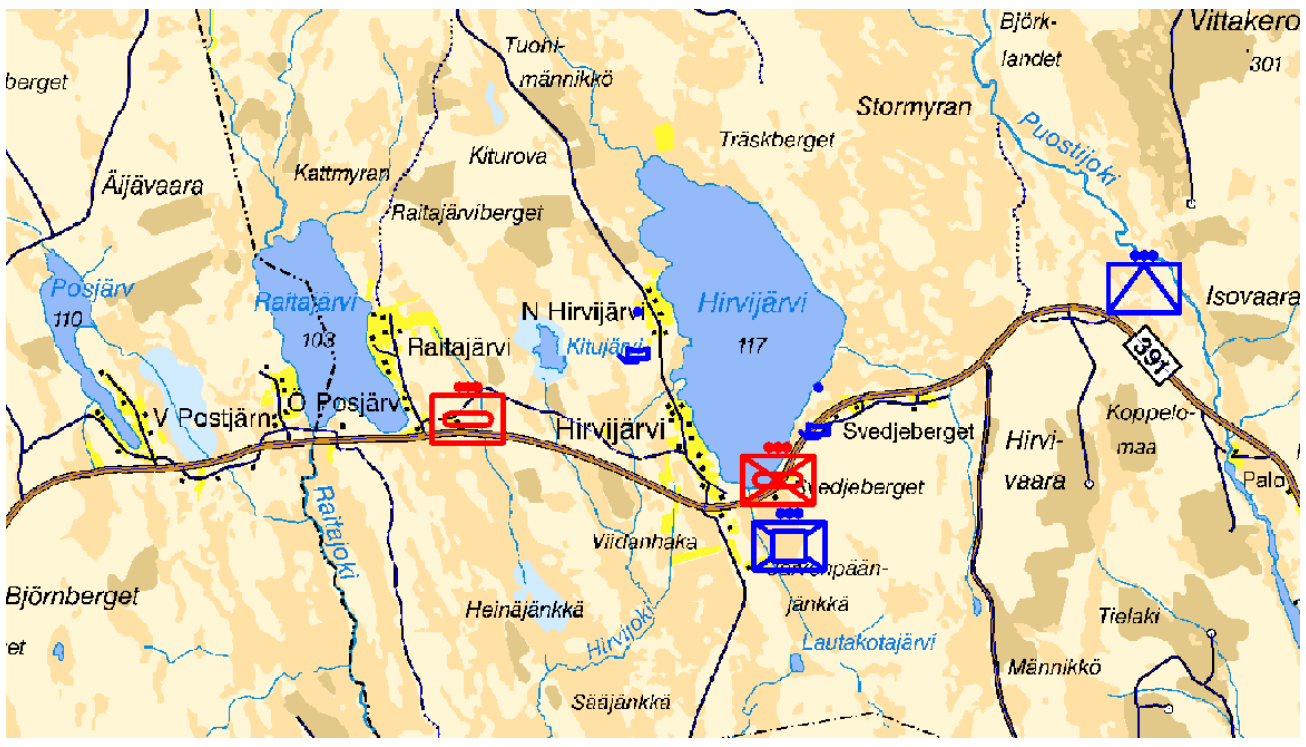

Figure 17: All platoons in IS Mark.

### 5.2 *Limitations and possibilities*

There is an inherent problem in the methods described above: what if there are several different ways of clustering objects that are roughly equal in value? Our experience shows that this is more or less the normal state of affairs and not the exception. Choosing one of these and presenting it as the truth, or as the description of the situation that is most likely to be true given the evidence, will be more or less arbitrary.

It is not a new problem: how should one present uncertain conclusions when there are rival conclusions to be drawn? The problem is compounded by the time critical nature of the endeavour. A user will often not have time for a deep analysis of the material available: if he did he would probably have limited use of a method for automatic force aggregation to begin with. Thus one cannot get around the problem by saying that no automatic decision about what hypotheses to present should be made at all. Even if there is a risk that the wrong conclusion is drawn, this risk must be taken.

## 6 Conclusions

In this papers we have shown that it is possible to perform automatic force aggregation for the lowest levels (intelligence reports to vehicles, and vehicles to platoons) by clustering and classification using only conflicts between elements.

For the future we aim to study higher levels in the hierarchy. We believe it will then be necessary to use both positive relations (reasons elements might belong together) as well as the negative conflicting relations used in this paper for clustering together with more advanced templates for classification.